\documentclass{article}

\usepackage[final]{nips_2016}

\usepackage[utf8]{inputenc} 
\usepackage[T1]{fontenc}    
\usepackage{hyperref}       
\usepackage{url}            
\usepackage{booktabs}       
\usepackage{amsfonts}       
\usepackage{nicefrac}       
\usepackage{microtype}      

\usepackage{algorithm}
\usepackage{algorithmic}
\usepackage{amsmath,amsfonts,amssymb}
\usepackage{graphicx}
\usepackage{subfigure} 

\newtheorem{theorem}{Theorem}

\newenvironment{proof}[1][Proof]{\begin{trivlist}
\item[\hskip \labelsep {\bfseries #1}]}{\end{trivlist}}

\newcommand{\bs}{\boldsymbol}
\newcommand{\w}{{\boldsymbol \theta}}
  
\newcommand{\x}{{\boldsymbol \phi}}

\usepackage{xspace}

\title{Effective Multi-step Temporal-Difference Learning \\for Non-Linear Function Approximation}

\author{
  Harm van Seijen \\
  \\
  RLAI Lab\\
  University of Alberta\\
   \\
  Maluuba Research\\
  2000 Peel Street, Montreal, QC\\
  Canada, H3A 2W5\\
  \\
  \texttt{harm.vanseijen@maluuba.com} \\
}

\begin{document}

\maketitle

\begin{abstract} 
Multi-step temporal-difference (TD) learning, where the update targets contain information from multiple time steps ahead, is one of the most popular forms of TD learning for linear function approximation. The reason is that multi-step methods often yield substantially better performance than their single-step counter-parts, due to a lower bias of the update targets. For non-linear function approximation, however, single-step methods appear to be the norm. Part of the reason could be that on many domains the popular multi-step methods TD($\lambda$) and Sarsa($\lambda$) do not perform well when combined with non-linear function approximation. In particular, they are very susceptible to divergence of value estimates. In this paper, we identify the reason behind this. Furthermore, based on our analysis, we propose a new multi-step TD method for non-linear function approximation that addresses this issue. We confirm the effectiveness of our method using two benchmark tasks with neural networks as function approximation.
\end{abstract}

\section{Introduction}

Multi-step update targets play an important role in TD learning \citep{sutton:ml88} and reinforcement learning \citep{sutton:book98, szepesvari:book09}. The core concept behind TD learning is to bootstrap the value of one state (or state-action pair) from the value of another state (or state-action pair). With one-step update targets the state that is bootstrapped from lies one time step in the future; with multi-step update targets bootstrapping occurs with respect to values of states that lie further in the future. Controlling from which states bootstrapping occurs is important, because it affects the fundamental trade-off between bias and variance of updates. The trade-off that produces the best performance is different from domain to domain, but for most domains the best trade-off lies somewhere in between a one-step update target (high bias, but low variance) and an update with the full return (unbiased, but high variance). This has made TD($\lambda$), where the trade-off between variance and bias of the update target can be controlled by the parameter $\lambda$, one of the most popular TD methods in linear function approximation.

While TD($\lambda$) and its control variant Sarsa($\lambda$) are very popular in the case of linear function approximation, when non-linear function approximation is used to represent the value function single-step methods are the norm. A reason could be that in many domains with non-linear function approximation TD($\lambda$) does not perform particularly well. In particular, it is very susceptible to divergence of values.  We argue that the underlying reasons for this instability are not unique to non-linear function approximation; it is a more general phenomenon of traditional TD($\lambda$). However, the issues are more prominent when non-linear function approximation is used for two reasons. First, for table lookup or linear function approximation with binary features, an alternative version of TD($\lambda$) is available (TD($\lambda$) with replacing traces) that is less sensitive to divergence \citep{singh:ml96}. Second, value blow-ups occur especially in domains where the same feature is active (i.e., has a value $\neq 0$) for many subsequent time steps \citep{vanseijen:arxiv15b}. This is something that occurs often with non-linear function approximation, because features are typically more general in this setting and can be active over a large part of the state-space.

We show that the susceptibility of TD($\lambda$) to divergence stems from a deviation of TD($\lambda$) from the general TD update rule based on gradient descent that is formalized by its forward view. Unfortunately, while the forward view is less susceptible to divergence, it is expensive to implement (both the computation time per step and required memory grow over time), making it not a practical alternative to TD($\lambda$). To address this, we present an alternative version of TD($\lambda$), which we call forward TD($\lambda$), that implements the gradient-descent-based update rule exactly and is computationally efficient as well. The price that is payed to achieve this is that updates occur with a delay. However, we show empirically that the advantages of having an exact implementation of the gradient-descent-based update rule substantially outweigh the disadvantages of having a delay in the updates.

\section{Related Work}

This work is related to true online temporal-difference learning \citep{vanseijen:icml14, vanseijen:arxiv15b}. The non-linear, online $\lambda$-return algorithm presented in Section \ref{sec:analysis} is a direct extension of the linear, online $\lambda$-return algorithm that underlies true online TD($\lambda$). In the linear case, the computationally inefficient forward view equations can be rewritten in computationally efficient backward view equations, yielding the true online TD($\lambda$) algorithm. Unfortunately, this is not possible in the non-linear case, because the derivation of the true online equations makes use of the fact that the gradient with respect to the value function is independent of the weight vector, which does not hold in the case of non-linear function approximation.  

Forward TD($\lambda$) is similar to a method introduced by \citet{cichosz:jair95}. Specifically, Cichosz's method is based on the same update target as forward TD($\lambda$). Interestingly, Cichosz presents his method in the context of linear function approximation as a computationally efficient alternative to traditional TD($\lambda$). While we focus primarily on sample efficiency in the non-linear setting, like Cichosz's method, forward TD($\lambda$) also has computational advantages. In fact, forward TD($\lambda$) is more efficient than Cichosz's method. Forward TD($\lambda$) has the same computation-time complexity as TD(0);  by contrast, the computation-time of Cichosz's method depends on $K$.

\section{Background}

Our problem setting is that of a \emph{Markov decision processes} (MDP), which can be described as a 5-tuple of the form $\langle \mathcal{S}, \mathcal{A}, p, r, \gamma \rangle$, consisting of $\mathcal{S}$, the set of all states; $\mathcal{A}$, the set of all actions; $p(s'|s,a)$, the transition probability function, giving for each state $s \in \mathcal{S}$ and action  $a \in \mathcal{A}$  the probability of a transition to state $s' \in \mathcal{S}$ at the next step;  $r(s,a,s')$, the reward function, 
giving the expected reward for a transition from $(s,a)$ to $s'$. $\gamma$ is the discount factor, specifying how future rewards are weighted with respect to the immediate reward. An MDP can contain terminal states, which terminate an episode. Mathematically, a terminal state can be interpreted as a state with a single action that results in a reward of 0 and  a transition to itself.

The return at time $t$ is defined as the discounted sum of rewards, observed after $t$:
\begin{displaymath}
G_t = R_{t+1} + \gamma\,R_{t+2} + \gamma^2\,R_{t+3}+... = \sum_{i=1}^\infty\,\gamma^{i-1}\, R_{t+i}\thinspace,
\end{displaymath}
where $R_{t+1}$ is the reward received after taking action $A_t$ in state $S_t$. 

Actions are taken at discrete time steps $t = 0,1,2,...$ according to a \emph{policy} $\pi: \mathcal{S} \times \mathcal{A} \rightarrow [0,1]$, which defines for each action the selection probability conditioned on the state. Each policy $\pi$ has a corresponding state-value function $v_{\pi}(s)$, which maps each state $s \in \mathcal{S}$ to the expected value of the return $G_t$ from that state, when following policy $\pi$:
$$v_{\pi}(s) = \mathbb{E}\{ G_t \,|\, S_t = s, \pi \}\thinspace.$$
The value of a terminal state is (by definition) 0.

Temporal-Difference (TD) learning aims to learn the state-value function using a strategy based on stochastic gradient descent \citep{bertsekas:book95}. Let $\hat V(s | \w)$ be an estimate of $v_\pi(s)$ given the weight vector $\w \in \mathbb{R}^n$.  Then, the general form of the TD update is:
\begin{equation}
\w_{t+1}  = \w_t + \alpha \Big( U_t - \hat V(S_t | \w_t) \Big) \nabla_\theta \hat V(S_t | \w_t)\,,
\label{eq:non-linear update}
\end{equation}
where $\alpha > 0$ is the step-size parameter, $U_t$ is the update target, and $\nabla_\theta \hat V$ is the gradient of $\hat V$ with respect to the weight vector $\w$. The update target $U_t$ is some estimate of the value $v_\pi (S_t)$. A simple example is the TD(0) update target, which uses the estimate of the next state to bootstrap from:
$$U_t = R_{t+1} + \gamma \hat V(S_{t+1}|\w_t)\,.$$

The update equations for TD($\lambda$) are:
\begin{eqnarray*}
\delta_t &=& R_{t+1} + \gamma \hat V(S_{t+1}| \w_t)   - \hat V(S_{t}| \w_t) \\
{\bs e}_t &=& \gamma\lambda {\bs e}_{t-1} +  \nabla_\w  \hat V(S_{t}| \w_t) \\
\w_{t+1} &=&  \w_t + \alpha \delta_t\,{\bs e}_{t}
\end{eqnarray*}
where ${\bs e}_t$ is called the \emph{eligibility-trace vector}. 
While these updates appear to deviate from the gradient-descent-based update rule given in (\ref{eq:non-linear update}), there is a close connection with this update rule. In the next section, we go deeper into the details of this relation.

\section{Analysis of TD($\lambda$)}
\label{sec:analysis}

That TD($\lambda$) is a multi-step method is not immediately obvious, because its update equations are different in form than (\ref{eq:non-linear update}), making it hard to specify what the update target is. That TD($\lambda$) is a multi-step method becomes clear from the fact that the weights computed by TD($\lambda$) are similar to those computed by a different algorithm that does have a well-defined multi-step update target, called the $\lambda$-return algorithm. The $\lambda$-return algorithm is also referred to as the forward view of TD($\lambda$). While the traditional $\lambda$-return algorithm is similar to TD($\lambda$) only at the end of an episode \citep{sutton:book98, bertsekas:book96}, below we specify a more general version that is similar to TD($\lambda$) at \emph{all time steps}.

We define the $\lambda$-return for time step $t$ with horizon $h \geq t+1$ as follows:
\begin{equation}
G^{\lambda|h}_t := (1-\lambda) \sum_{n=1}^{h-t-1}  \lambda^{n-1} G_t^{(n)} + \lambda^{h-t-1} G_t^{(h-t)}\,
\label{eq:interim lambda return1}
\end{equation}
where $G_t^{(n)}$ is the $n$-step return, defined as:
\begin{displaymath}
G_t^{\,(n)} := \sum_{k=1}^n \gamma^{k-1} R_{t+k} + \gamma^n\, \hat V(S_{t+n}| \w_{t+n-1}).
\end{displaymath}
Note that $G_t^{\lambda|h}$ uses information only up to the horizon $h$. We define $\w_t$ as the result of a sequence of updates of the form (\ref{eq:non-linear update}), based on states $S_0, \dots, S_{t-1}$ and update targets $G_0^{\lambda|t}, \dots, G_{t-1}^{\lambda|t}$, respectively.  Formally, we define $\w_t := \w_t^t$, with $\w_t^t$ incrementally defined by:\footnote{Note that the sequence of updates is different for each time step, due to the different horizons, requiring the double indices for the weight vectors.}
\begin{equation}
\w_{k+1}^t  := \w_k^t + \alpha \Big( G_k^{\lambda|t} - \hat V(S_k | \w_k^t) \Big) \nabla_\theta \hat V(S_k | \w_k^t), \qquad\mbox{for } 0 \leq k < t\,.
\label{eq:interim update}
\end{equation}
with $\w_0^t := \w_0$ for all $t$ and $\w_0$ being the weight vector at the start of the episode. We call the algorithm that implements these updates the \emph{online $\lambda$-return algorithm}. 
Furthermore, we define the \emph{offline $\lambda$-return algorithm} as the algorithm that performs (\ref{eq:interim update}) only at the end of an episode. That is, $\w_t := \w_0$ for $0 \leq t < T$,  with $T$ the time step of termination, while $\w_T: = \w_T^T$, with $\w_T^T$ defined incrementally by (\ref{eq:interim update}).

\begin{figure}[thb]
\begin{center}
\includegraphics[width=9cm]{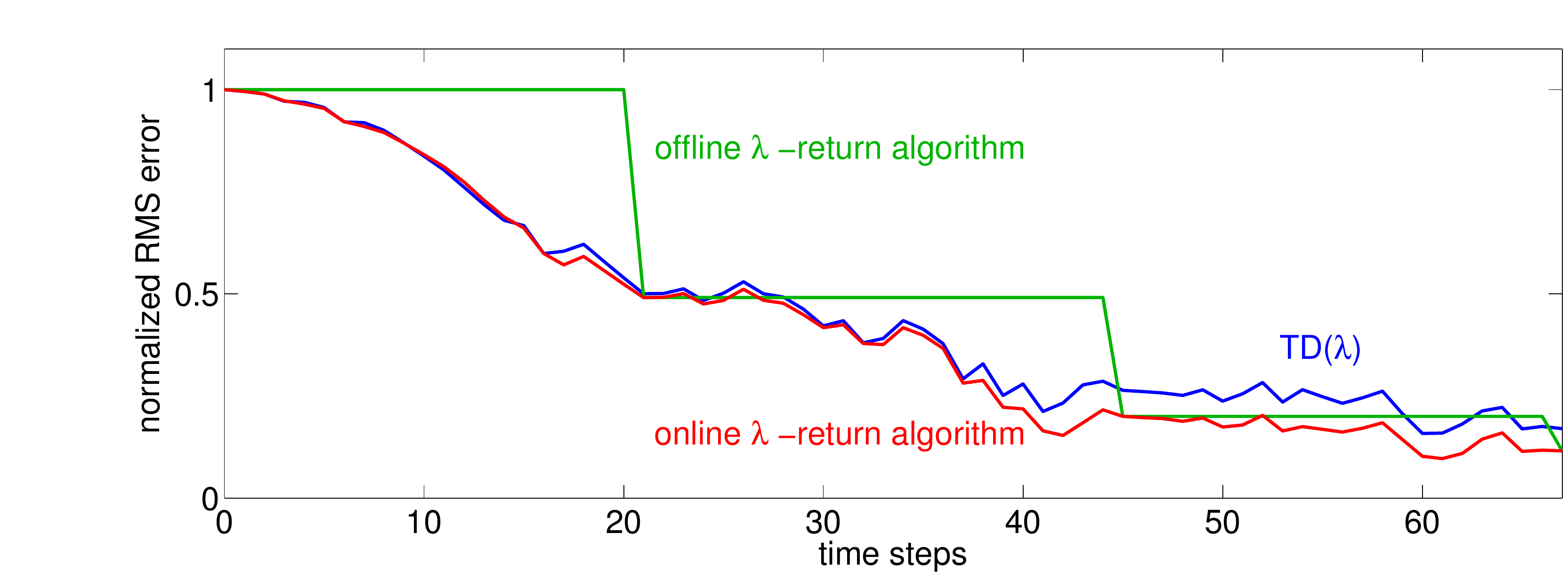}
\caption{RMS error as function of time, for the first 3 episodes of a random walk task, for $\lambda = 1$ and $\alpha = 0.2$. The error shown is the RMS error over all states, normalized by the initial RMS error.} 
\label{fig:offline vs online}
\end{center}
\end{figure}

Figure \ref{fig:offline vs online} illustrates the difference between the online and offline $\lambda$-return algorithm and TD($\lambda$), by showing the RMS error on a random walk task. The task consists of 10 states laid out in a row plus a terminal state on the left. Each state transitions with 70\% probability to its left neighbour and with 30\% probability to its right neighbour (or to itself in case of the right-most state). All rewards are 1, and $\gamma = 1$. The right-most state is the initial state. 

The theorem below states that for appropriately small step-sizes TD($\lambda$) behaves like the online $\lambda$-return algorithm. We provide the proof for the theorem in Appendix \ref{sec:proof}. The theorem uses the term $\Delta_i^t$, which we define as:
$$\Delta_i^t := \big(\bar G_i^{\lambda|t} - \hat V(S_i | \w_0)\big)\nabla_\w \hat V(S_i | \w_0)\,,$$ with $\bar G_i^{\lambda|t}$ the interim $\lambda$-return for state $S_i$ with  horizon $t$ that uses $\w_0$ for all value evaluations. Note that $\Delta_i^t$ is independent of the step-size.

\begin{theorem}
Let $\w_0$ be the initial weight vector, $\w_{t}^{td}$ be the weight vector at time $t$ computed by TD($\lambda$), and $\w_{t}^{\lambda}$ be the weight vector at time $t$ computed by the online $\lambda$-return algorithm.
Furthermore, assume that $\nabla_\w \hat V$ is well-defined and continuous everywhere and that 
$\sum_{i=0}^{t-1} \Delta_i^t \neq {\boldsymbol 0}$.
Then, for all time steps $t$:
$$\frac{|| \w_{t}^{td}  - \w_{t}^{\lambda}  ||}{|| \w_{t}^{td}  - \w_0  ||} \rightarrow 0 \qquad\mbox{as $\,\,\,\alpha \rightarrow 0$}.$$
\end{theorem}

\begin{figure}[thb]
  \begin{minipage}[c]{0.5\textwidth}
  \hspace{1cm}
    \includegraphics[width=0.5\textwidth]{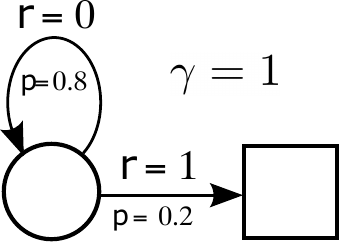}
  \end{minipage}\hfill
  \begin{minipage}[c]{0.6\textwidth}
   \includegraphics[width=0.6\textwidth]{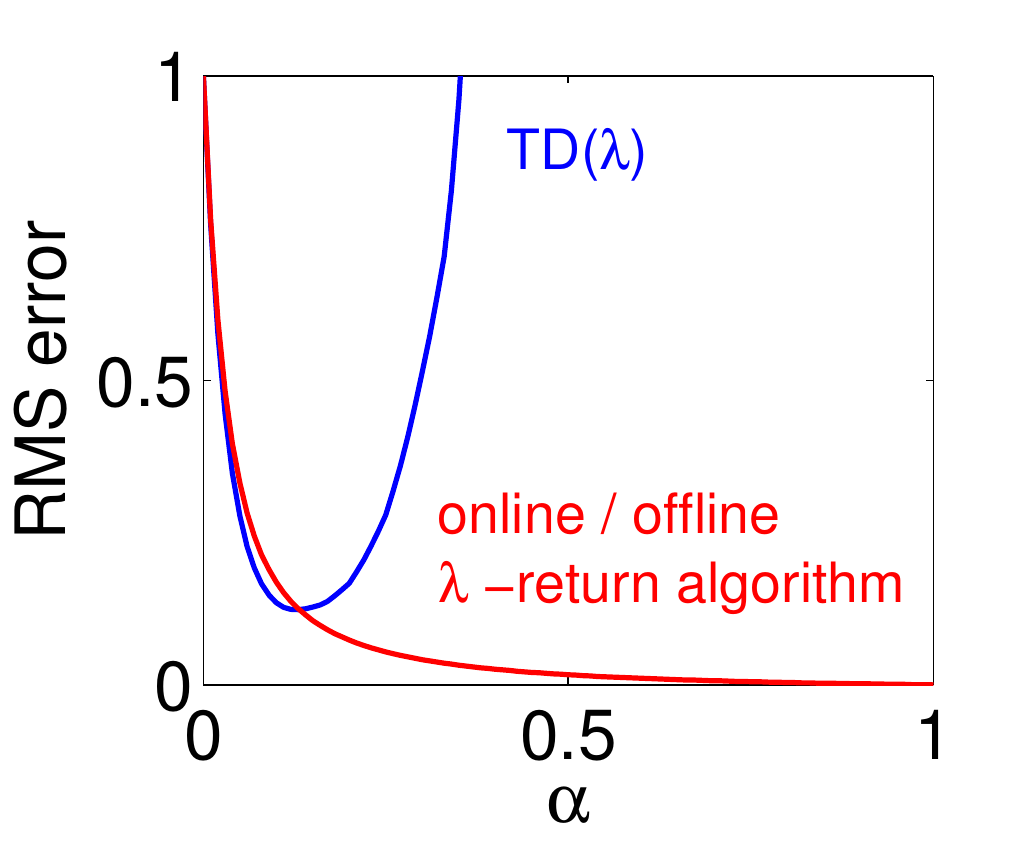}
  \end{minipage}
  \vspace{-0.2cm}
      \caption{{\it Left: } One-state example (the square indicates a terminal state). {\it Right: } The RMS error of the state value at the end of an episode, averaged over the first 10 episodes, for $\lambda = 1$.}
          \label{fig:one-state example}
\end{figure}

While TD($\lambda$) behaves for small step-size like the $\lambda$-return algorithm, in practise a small step-size often results in slow learning. Hence, higher step-sizes are desirable. Figure \ref{fig:offline vs online} suggests that for higher step-sizes, TD($\lambda$) has a disadvantage with respect to the online $\lambda$-return algorithm. We analyze why this is the case, using the one-state example shown in the left of Figure \ref{fig:one-state example}. 
\vspace{-0.1cm}

The right of Figure \ref{fig:one-state example} shows the RMS error over the first 10 episodes for different step-sizes and $\lambda = 1$. While for small step-sizes, TD($\lambda$) indeed behaves like the $\lambda$-return algorithm, for larger step-sizes the difference becomes huge. 

To understand the reason for the large difference in performance, we derive an analytical expression for the value at the end of an episode. First, we consider the $\lambda$-return algorithm. Because there is only one state involved, we indicate the value of this state simply by $\hat V$. The value at the end of an episode, $\hat V_T$, is equal to $\hat V_T^T$, resulting from the update sequence:
$$ \hat V_{k+1}^T = \hat V^T_k + \alpha ( G_k^{\lambda | T} - \hat V^T_k) \qquad \mbox{ for } 0 \leq k < T$$
By substitution, we can directly express $\hat V_T$ in terms of the initial value, $\hat V_0$, and the update targets:
$$ \hat V_T = (1-\alpha)^T \hat V_0 + \alpha (1-\alpha)^{T-1} G_0^{\lambda | T}  + \alpha (1-\alpha)^{T-2} G_1^{\lambda | T}  + \cdots + \alpha G_{T-1}^{\lambda | T} $$
Using that $G_k^{\lambda|T} =  1$ for all $k$, this can be written as a single pseudo-update:
\begin{equation}
\hat V_T  = \hat V_0 + \beta  (1 - \hat V_0)
\label{eq:VT}
\end{equation}
with $\beta = 1 - (1-\alpha)^T$. Note that a larger $\alpha$ or $T$ results in a larger $\beta$, but its value is bounded. Specifically, $ 0 \leq \alpha \leq 1 \Rightarrow  0 \leq \beta \leq 1$.

We now consider TD($\lambda$). The update at the end of an episode is  $\hat V_T = \hat V_{T-1} + \alpha e_{T-1} \delta_{T-1}$ .
In our example, $\delta_t = 0$ for $0 \leq t < T-1$, while $\delta_{T-1} = 1 - V_{T-1}$. Because $\delta_t$  is 0 for all time steps except the last, $V_{T-1} = V_0$.  Furthermore, $\nabla_\w \hat V$ reduces to 1 in our example, resulting in $e_{T-1}= T$. Substituting all this in the above equation also reduces it to pseudo-update (\ref{eq:VT}), but with $\beta  = \alpha T$. So for TD($\lambda$), $\beta$ can grow much larger than 1, causing divergence of values, even for $\alpha < 1$. This is the reason that TD($\lambda$) can be very sensitive to the step-size and it explains why the optimal step-size for TD($\lambda$) is much smaller than the optimal step-size for the $\lambda$-return algorithm in Figure \ref{eq:VT} ($\alpha \approx 0.15$  versus $\alpha = 1$, respectively). Moreover, because the variance on $\beta$ is higher for TD($\lambda$) the performance at optimal $\alpha$ of TD($\lambda$) is worse than the performance at optimal $\alpha$ for the $\lambda$-return algorithm. In Section \ref{sec:empirical}, we show empirically that the general behaviour of TD($\lambda$) shown in Figure \ref{fig:one-state example} also occurs in more complex domains.  

While the online $\lambda$-return algorithm has clear advantages over TD($\lambda$), it is not a practical algorithm: the number of updates that need to be performed per time step grows over time, as well as the memory requirements. On the other hand, the offline $\lambda$-return algorithm is undesirable, because it performs no updates during an episode and cannot be applied to non-episodic tasks. In the next section, we present forward TD($\lambda$), a computationally efficient algorithm that forms a middle ground between the online and the offline $\lambda$-return algorithm.

\section{Forward TD($\lambda$)}

The online $\lambda$-return algorithm uses update targets that grow with the data horizon. This has the advantage that updates can be performed immediately, but also causes the computation time per time step to grow over time. In this section, we present a computationally efficient method that performs updates using a $\lambda$-return with a horizon that lies a fixed number of time steps in the future: $G_t^{\lambda|t+K}$ with $K \in \{1, 2, \dots \}$. We refer to this update target as the $K$-bounded $\lambda$-return.

A consequence of using update target $G_t^{\lambda|t+K}$ with fixed $K$ is that during the first $K-1$ time steps no updates occur. In other words, $\w_t  := \w_0$ for $1 \leq t < K$. The weights $\w_K$ through $\w_{T-1}$ are defined as follows:
$$\w_{t+K} := \w_{t+K-1} + \alpha \Big( G_{t}^{\lambda|t+K} - \hat V(S_{t} | \w_{t +K-1}) \Big) \nabla_\theta \hat V(S_{t} | \w_{t+K-1})\,,  \qquad\mbox{ for } \,\,0 \leq t < T - K.$$
At the end of an episode $K$ updates occur. Following the convention of the double indices when multiple updates occur at a single time step, we define $\w_T := \w^T_K$, with $\w^T_K$ defined incrementally by:
$$\w_{k+1}^T := \w_{k}^T + \alpha \Big( G_{T-K+k}^{\lambda|T} - \hat V(S_{T-K+k} | \w_{k}^T) \Big) \nabla_\theta \hat V(S_{T-K+k} | \w_{k}^T)  \qquad\mbox{ for } \,\,0 \leq k < K\,,$$
with $\w_0^T := \w_{T-1}$. 

The question of how to set $K$ involves a trade-off. On the one hand, larger values of $K$ bring the end-of-episode weights closer to those of the $\lambda$-return algorithm; on the other hand, smaller values of $K$ result in a shorter delay of updates. In general, $K$ should be set in such a way that $G_t^{\lambda|t+K}$ is an accurate estimate of $G_t^{\lambda|T}$, while not being unnecessary large. How accurately  $G_t^{\lambda|t+K}$ approximates $G_t^{\lambda|T}$ depends on the value $\gamma\lambda$, because the contribution of a reward to the $K$-bounded $\lambda$-return reduces exponentially with $\gamma\lambda$ (we will show this below). While the immediate reward has a contribution of 1, the contribution of a reward $K$ time steps in the future is only $(\gamma\lambda)^K$. Hence, a sensible strategy for setting $K$ is to find the smallest value of $K$ that still ensures that the value $(\gamma\lambda)^K$ is smaller than some fraction $\eta$.  This value can be computed as follows:
\begin{equation}
K =  ceil\big(  log(\eta) / log(\gamma\lambda) \big)\,,
\label{eq:K value}
\end{equation}
where $ceil(\cdot)$ rounds up to the nearest integer. 
Note that for $\gamma\lambda < \eta $ ,  $ K = 1$. The value $K=1$ is special because $G_t^{\lambda|t+1}$ reduces to the TD(0) update target, independent of the value of $\lambda$. Furthermore, there is no delay in updates. Hence, forward TD($\lambda$) behaves exactly like $TD(0)$ in this case. For $\gamma\lambda = 1$, no finite value of $K$ can ensure that an accurate estimate of $G_t^{\lambda|T}$ is obtained. The only way to resolve this is to postpone all updates to the end of an episode (which can be interpreted as $K = \infty$). In this case, the performance of forward TD($\lambda$) is equal to that of the offline $\lambda$-return algorithm.

Next, we discuss how forward TD($\lambda$) can be implemented efficiently. Our implementation is based on two ways of computing the $K$-bounded $\lambda$-return. We derive the underlying equations in Appendix \ref{sec:bounded lambda-return}.  The first way is based on the equation:
\begin{equation}
G^{\lambda | h+1}_t  = G^{\lambda | h}_t   + (\gamma\lambda)^{h-t} \delta'_{h}\,,  \qquad\mbox{ for } h \geq t+1\,,
\label{eq:Glambda h+1}
\end{equation}
with
$$\delta'_h := R_{h+1} + \gamma  \hat V(S_{h+1} | \w_{h})  - \hat V(S_{h} | \w_{h-1})  \,.$$
Note that $\delta'_i$ differs from $\delta_i$ in the index of the weight vector used for the value of $S_i$.
Using (\ref{eq:Glambda h+1}) incrementally, $G_t^{t+K}$ can be computed, starting from $G_t^{\lambda| t+1} = R_{t+1} + \gamma \hat V(S_{t+1}| \w_t)$, in $K-1$ updates.

The second way is based on the equation:
\begin{equation}
G^{\lambda | h}_{t+1} = (G^{\lambda | h}_t - \rho_t )/\gamma\lambda\,, \qquad\mbox{ for } h \geq t+2\,,
\label{eq:update with rho}
\end{equation}
with
$$ \rho_t = R_{t+1} + \gamma(1-\lambda)\, \hat V(S_{t+1} | \w_t)\,.$$
This equation can be used to compute $G_{t+1}^{t+K}$ from $G_{t}^{t+K}$. Performing one more update using (\ref{eq:Glambda h+1}) results in the $K$-bounded $\lambda$-return for time step $t+1$:  $G_{t+1}^{\,t+1+K}$. This way of computing the $K$-bounded $\lambda$-return requires only two updates (for any value of $K$).

In theory, the $K$-bounded $\lambda$-return has to be computed incrementally from scratch (using Equation \ref{eq:Glambda h+1}) only for the initial state; for the other states it can be computed efficiently using only 2 updates. Unfortunately, this approach does not work well in practise. The reason is that tiny rounding errors that occur on any computer get blown up by dividing by $\gamma\lambda$ over and over again. For example, consider $\gamma\lambda = 0.5$. Then, rounding errors in the $K$-bounded $\lambda$-return at time $t$ will be blown up by a factor  $(1 / \gamma\lambda)^{100} = 2^{100}$ at time $t+100$. Fortunately, we can avoid these blow-ups in an elegant way, by recomputing the $K$-bounded $\lambda$-return from scratch every $K$ time steps. This ensures that rounding errors will never grow by a factor larger than $(1/\gamma\lambda)^{K}$. Moreover, as we argued in the previous subsection, $K$ is set in such a way that the value $\gamma\lambda^K$ is just slightly smaller than the hyper-parameter $\eta$. Hence, rounding errors will not grow by a factor larger than approximately $1/\eta$. Because $\eta$ will typically be set to $0.01$ or larger (smaller values of $\eta$ will result in longer update delays, which is undesirable), no issues with rounding error blow-ups will occur.

We now analyze the computational complexity of forward Sarsa($\lambda$). For reference purposes, the pseudocode for implementing forward TD($\lambda$) in provided in Algorithm \ref{al:forward TD(lambda)}.  First, we look at computation time. Between time step $K$ and the end of an episode, exactly one state-value evaluation and one state-value update occur.  All other computations have $\mathcal{O}(1)$ cost. At the end of the episode an additional $K-1$ value updates occur, so there is a spike in computation at the end of an episode, but because during the first $K-1$ time steps of an episode no updates occur, on average the algorithm still performs only one value update and one value evaluation per time step. This is the same as for TD(0). Hence, forward TD($\lambda$) is very efficient from a computation time perspective.
In terms of memory, forward TD($\lambda$) requires the storage of the $K$ most recent feature vectors. So, if $n$ is the number of features, forward TD($\lambda$) requires additional memory of size $\mathcal{O}(nK)$ over TD(0) (note that forward TD($\lambda$) does not require storage of an eligiblity-trace vector). If $n$ is large and memory is scarce, $K$ can be bounded by some value $K_{\max}$ to deal with this.
\vspace{-0.2cm}

\begin{algorithm}[!thb]
\small
\begin{algorithmic}[0]
\STATE {\bf INPUT:} $\alpha, \lambda, \gamma, \w_{init},  \eta , K_{max}$ \it{(optional)}
\STATE $\w \leftarrow \w_{init}$
\STATE If $\,\,\gamma\lambda > 0\,\,$  then:  $\,\,K =  ceil\big(  \log(\eta) / \log(\gamma\lambda) \big)\,\,$,  else: $\,\,\,K = 1\,\,\,$
\STATE $K = \min (K_{max}, K)$ \qquad \it{(optional)}
\STATE $c_{final} \leftarrow (\gamma\lambda)^{K-1}$
\STATE Loop (over episodes):
\STATE \quad $\mathcal{F} \leftarrow \emptyset$   \qquad// $\mathcal{F}$ is a FIFO queue (max length: $K$)
\STATE \quad $U_{sync} \leftarrow 0; \,\,\,  i \leftarrow 0; \,\,\, c \leftarrow 1; \,\,\,  V_{current} \leftarrow 0; \,\,\, ready \leftarrow false $
\STATE \quad obtain initial state $S$		\qquad  // or $\x(S)$
\STATE \quad While $S$ is not terminal, do:
\STATE \quad\qquad observe reward $R$ and next state $S'$
\STATE \quad\qquad If $S'$ is terminal:  \,\,$V_{next} \leftarrow 0 $\,\,, else:  \,\,$V_{next} \leftarrow \hat V(S' | \w)$
\STATE \quad\qquad  $\rho \leftarrow R + \gamma (1-\lambda) V_{next}$
\STATE \quad\qquad push tuple $\langle S, \rho \rangle$ on $\mathcal{F}$  	\qquad  // or \,$\langle \x(S), \rho \rangle$
\STATE \quad\qquad $\delta' \leftarrow R + \gamma V_{next} - V_{current}$
\STATE \quad\qquad$V_{current} \leftarrow V_{next}$
\STATE \quad\qquad If $i = K -1 $ :
\STATE \quad\qquad\qquad $U \leftarrow U_{sync}$
\STATE \quad\qquad\qquad $U_{sync} \leftarrow V_{current}\,;\,\,\,i \leftarrow 0;\,\,\, c \leftarrow 1; \,\,\, ready \leftarrow true$
\STATE \quad\qquad Else:
\STATE \quad\qquad\qquad $U_{sync}  \leftarrow U_{sync}  + c \cdot \delta' $
\STATE \quad\qquad\qquad $ i \leftarrow i+1;\,\,\,c \leftarrow \gamma\lambda\cdot c$
\STATE \quad\qquad If $ready$ :
\STATE \quad\qquad\qquad $U \leftarrow U +  c_{final}\cdot \delta'$   \qquad\qquad//   $G_t^{\lambda|t+K} \Leftarrow G_t^{\lambda|t+K-1}$
\STATE \quad\qquad\qquad  pop $\langle S_p, \rho_p \rangle$ from $\mathcal{F}$
\STATE \quad\qquad\qquad  update $\w$ using $S_p$ and  $U$
\STATE \quad\qquad\qquad If $\,\, K \neq 1: \,\, U \leftarrow  \big(U - \rho_p \big) / (\gamma\lambda) $   \qquad\qquad//   $G_{t+1}^{\lambda|t + K} \Leftarrow G_t^{\lambda| t + K}$
\STATE \quad\qquad $S \leftarrow S'$
\STATE \quad  If $ready = false$:   \,\,\,$U \leftarrow U_{sync}$
\STATE \quad While  $\mathcal{F}$ not empty:
\STATE \quad\qquad  pop $\langle S_p, \rho_p \rangle$ from $\mathcal{F}$
\STATE \quad\qquad  update $\w$ using $S_p$ and  $U$
\STATE \quad\qquad   If $\,\, K \neq 1:\,\,  U \leftarrow  (U -   \rho_p) / \gamma\lambda$
\caption{forward TD($\lambda$)}
\label{al:forward TD(lambda)}
\end{algorithmic}
\end{algorithm}

\section{Empirical Comparisons}
\label{sec:empirical}

\begin{figure}[thb]
  \begin{minipage}[c]{0.5\textwidth}
  \hspace{1cm}
    \includegraphics[width=0.8\textwidth]{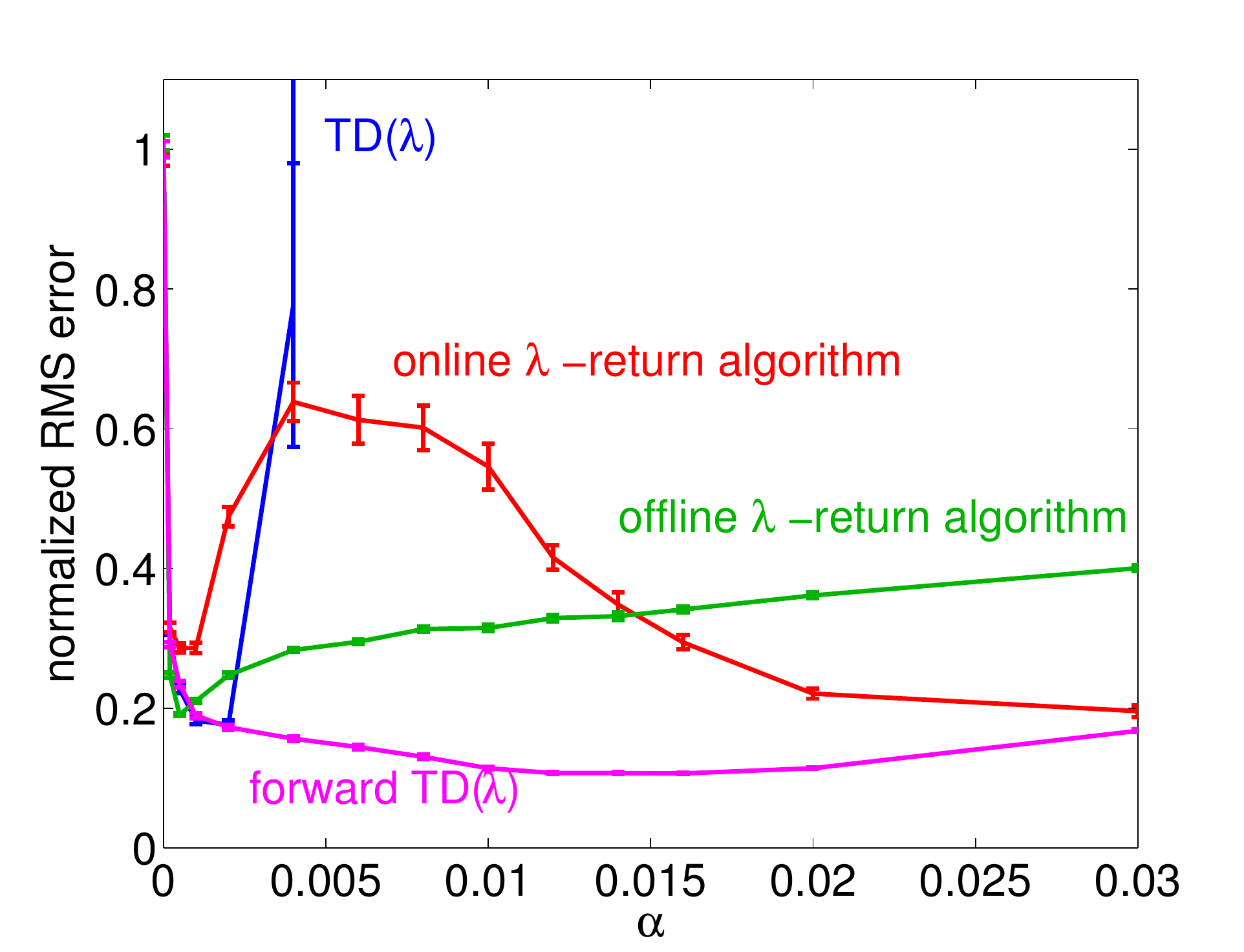}
  \end{minipage}\hfill
  \begin{minipage}[c]{0.5\textwidth}
   \includegraphics[width=0.8\textwidth]{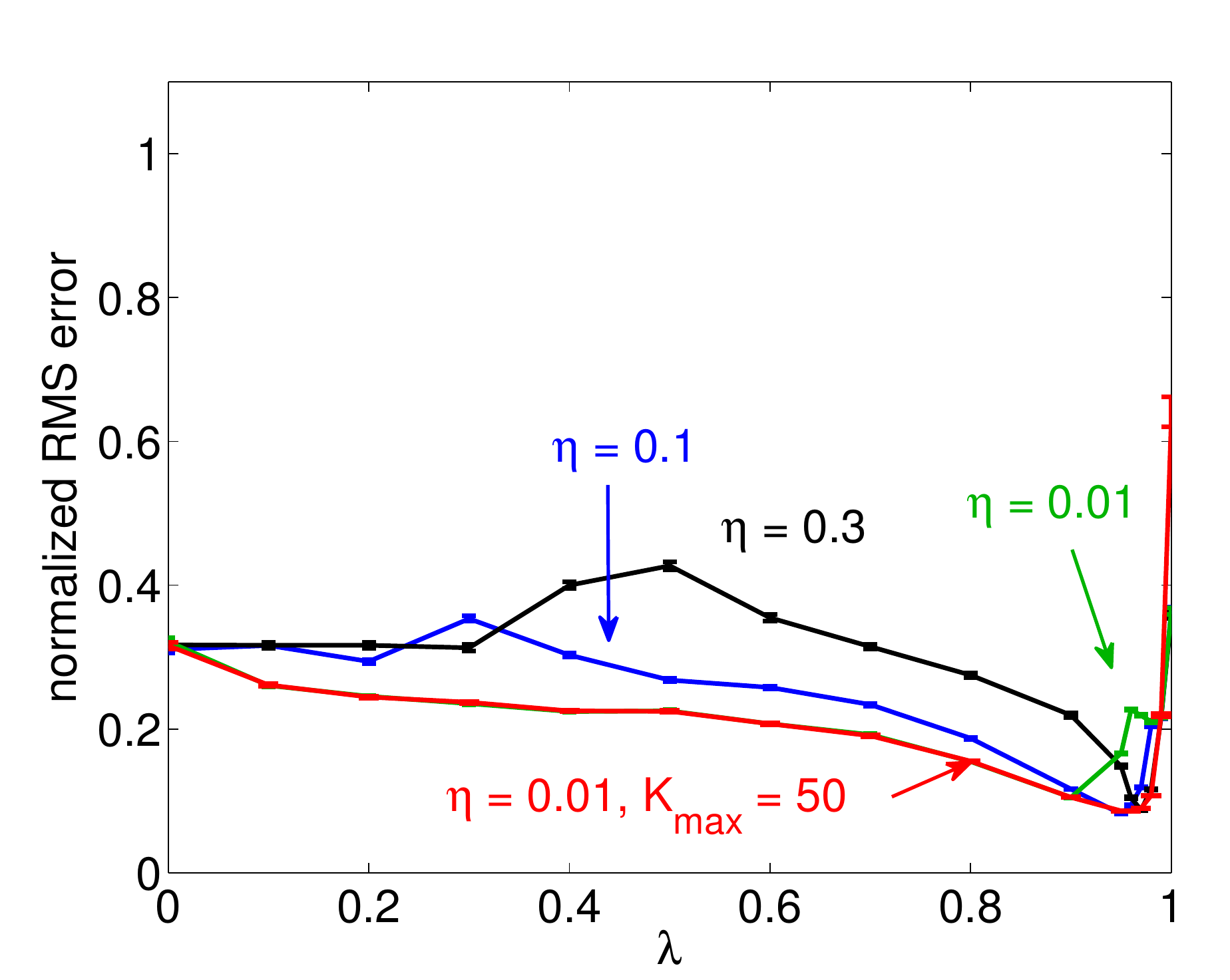}
  \end{minipage}
  \vspace{-0.2cm}
      \caption{RMS error averaged over the first 50 episodes of the mountain car evaluation task, normalized by the initial RMS error. {\it Left: } RMS error for different methods at $\lambda = 0.9$. {\it Right: } RMS error of forward TD($\lambda$) for different values of $\eta$ at $\alpha = 0.015$.} 
    \label{fig:mountain car eval}
\end{figure}

In our first experiment, we evaluate the performance of TD($\lambda$), forward TD($\lambda$) and the online/offline $\lambda$-return algorithm on the standard mountain car task \citep{sutton:book98}. The state-space consists of the position and velocity of the car, scaled
to numbers within the range [-1, 1]. The value function is approximated with a neural network that has the two state-variables as input, one output variable representing the state value, and a single hidden layer of 50 nodes in between. The backpropagation algorithm is used for obtaining the derivative of the value function with respect to the weights (in a similar way as done by Tesauro, 1994). 
The evaluation policy is a near-optimal policy. 
All rewards are drawn from a normal distribution with mean -1 and standard deviation 2. We fixed $\lambda = 0.9$ and set $\eta  = 0.01$ and show the performance for different step-sizes. Our performance metric is the RMS error (over the state distribution induced by the policy) at the end of an episode, averaged over the first 50 episodes. The left graph of Figure \ref{fig:mountain car eval} shows the results. The results are averaged over 50 independent runs. TD($\lambda$) shows the same behaviour as in the one-state example (Figure \ref{fig:one-state example}). That is, the error quickly diverges. Surprisingly, forward TD($\lambda$) outperforms the online $\lambda$-return algorithm. That delaying updates results in better performance in this case is probably related to the reason that the DQN algorithm uses a separate target network that is updated in a delayed way \citep{mnih:nature15}. Most likely because it reduces instability.

For our second experiment, we compared the performance of forward TD($\lambda$) with $\eta \in \{0.01, 0.1, 0.3 \}$ and no maximum $K$ value, for $\alpha = 0.015$ and different $\lambda$ values. In addition, we tested $\eta = 0.01$ with $K_{max} = 50$. The experimental settings are the same as in the first experiment, except we average over 200 independent runs instead of 50. The right graph of Figure \ref{fig:mountain car eval} shows the results. This graph shows that the performance at optimal $\lambda$ is not really affected by $\eta$. Hence, in practise $\eta$ can just be fixed to some small value.

For our third and fourth experiment, we used control tasks. Here the goal is to improve the policy in order to maximize the return. To deal with these tasks, we used one neural network per action to represent the action-value and used $\epsilon$-greedy action selection. Effectively, this changes TD($\lambda$) into Sarsa($\lambda$) and forward TD($\lambda$) into forward Sarsa($\lambda$).

\begin{figure}[thb]
  \begin{minipage}[c]{0.5\textwidth}
  \hspace{1cm}
    \includegraphics[width=0.8\textwidth]{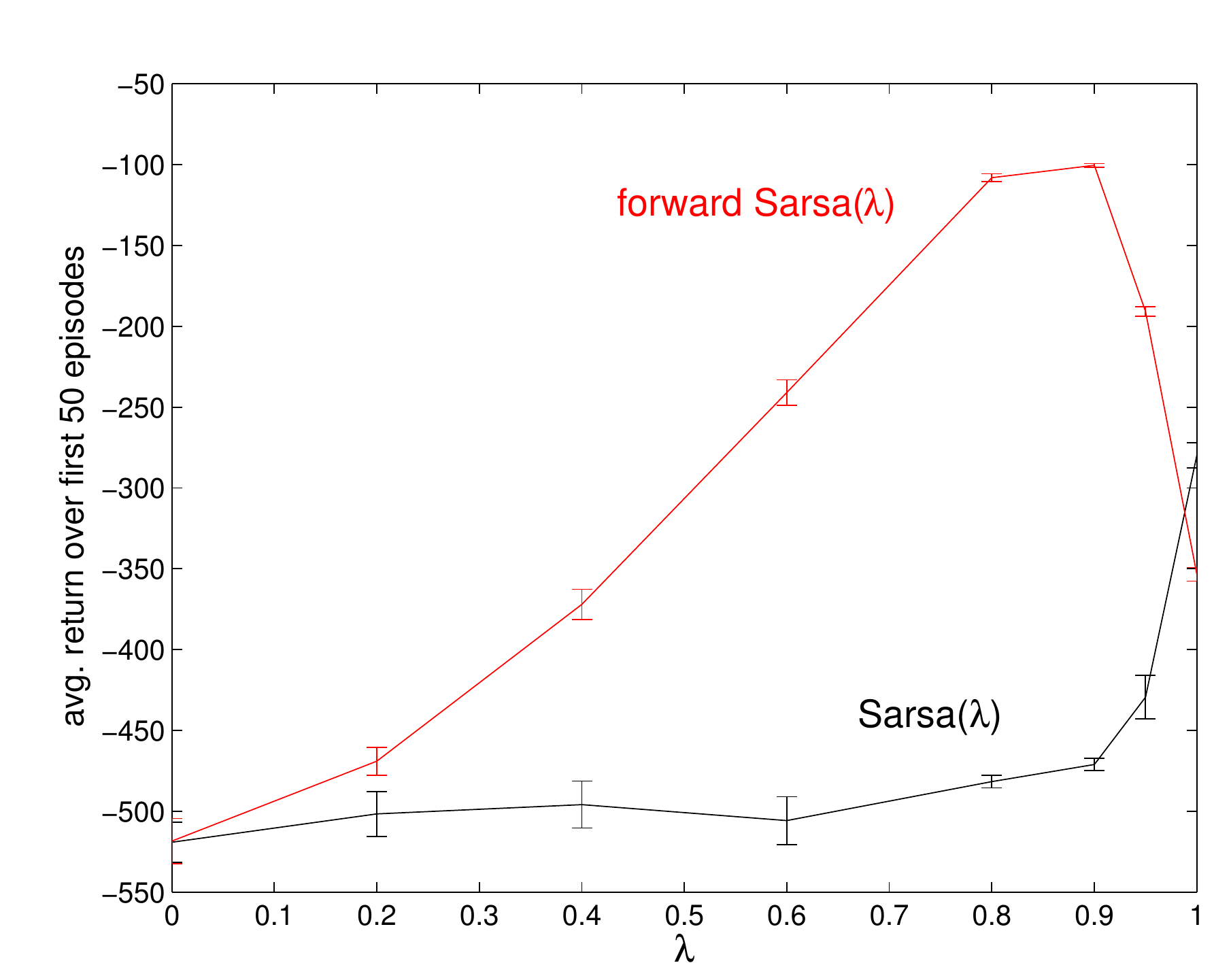}
  \end{minipage}\hfill
  \begin{minipage}[c]{0.5\textwidth}
   \includegraphics[width=0.8\textwidth]{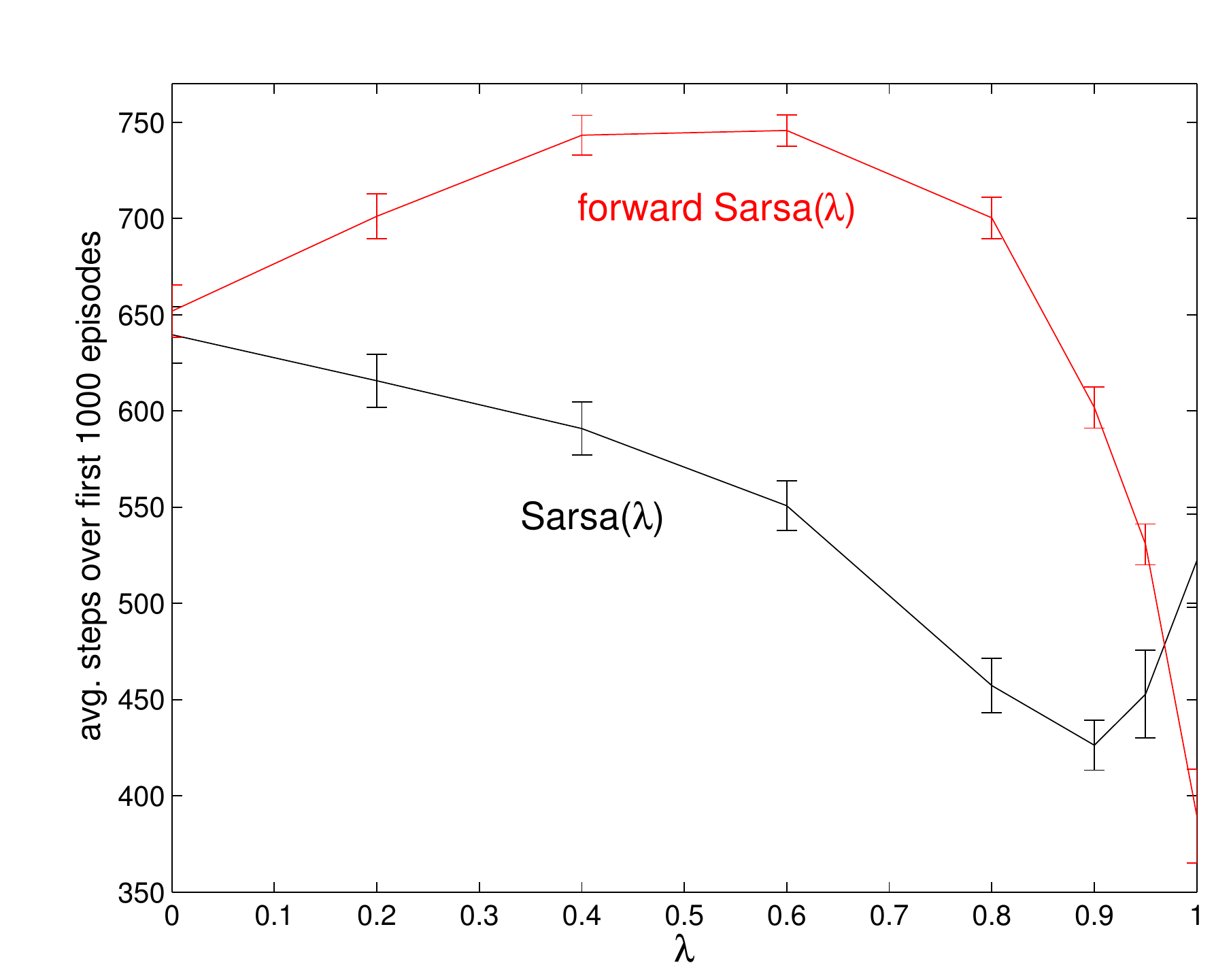}
  \end{minipage}
  \vspace{-0.2cm}
      \caption{Average return on two control tasks for different $\lambda$ and optimized $\alpha$ (and $\eta = 0.01$). {\it Left: } Mountain car task. {\it Right: } Cart-pole task. }
          \label{fig:control tasks}
\end{figure}

Our first control domain is the mountain car task, but now with deterministic rewards of -1. We compared the average return of Sarsa($\lambda$) and forward Sarsa($\lambda$) over the first 50 episodes for different $\lambda$. For each $\lambda$ and each method we optimized $\alpha$. We used $\eta = 0.01$ and $\epsilon = 0.05$. The left graph of Figure \ref{fig:control tasks} shows the results. Results are averaged over 200 independent runs. Forward Sarsa($\lambda$) outperforms Sarsa($\lambda$) for all $\lambda$ values, except for $\lambda = 1.0$. This can be explained by the fact that for $\lambda = 1$, all updates are delayed until the end of the episode for forward Sarsa($\lambda$), in contrast to the updates of Sarsa($\lambda$).

Our second control domain is the cart-pole benchmark task, in which a pole has to be balanced upright on a cart for as long as possible \citep{barto:smc83}. The state-space consists of the position and velocity of the cart, as well as the angle and angular velocity of the pole; there are only two actions: move left and move right. An episode ends when the angle of the pole deviates a certain number of degrees from its upright position or when the cart position exceeds certain bounds. We used $\epsilon$-greedy exploration with $\epsilon = 0.05$, and limited the episode length to 1000 steps. Again, $\eta = 0.01$. The networks we used for action-value estimation are the same as in the mountain car experiment (1 hidden layer consisting of 50 nodes), expect that each network now has four input nodes, corresponding with scaled versions of the four state-space parameters. We compared the average return over the first 1000 episodes for different $\lambda$ with optimized $\alpha$. The right graph of Figure \ref{fig:control tasks} shows the results, averaged over 200 independent runs. In this domain, higher values of $\lambda$ actually reduce the performance of Sarsa($\lambda$). By contrast, the optimal performance of forward Sarsa($\lambda$) is obtained around $\lambda = 0.6$ and is substantially higher than the performance of Sarsa(0). Overall, these results convincingly show that forward Sarsa($\lambda$) outperforms Sarsa($\lambda$), as predicted by our analysis.

\section{Conclusions}

We identified the reason why TD($\lambda$) often performs poorly on domains with non-linear function approximation. Deviations from the general TD update rule make TD($\lambda$) susceptible to divergence of value estimates and causes additional variance that reduces performance. While the $\lambda$-return algorithm implements the general update rule exactly, it is not a practical alternative, because its computation-time per step, as well as its memory requirements, are much more expensive. To address this, we presented a new method, called forward TD($\lambda$), that exactly implements the general update rule (like the $\lambda$-return algorithm), but  is also very efficient (like TD($\lambda$)). Specifically, its computation-time complexity is the same as that of TD(0). While forward TD($\lambda$) performs its updates with a delay, we have shown empirically that the performance increase due to exactly following the general update rule more than makes up for the performance decrease due to the update delays. In fact, one of our experiments suggests that the delay in updates could actually have a positive impact on the performance when non-linear function approximation is used. This surprising result is likely related to the same reason that DQN uses a separate target network that is updated in a delayed way and is an interesting topic for future research. 

\section*{Acknowledgements}
The author thanks Itamar Arel for discussions leading to the development of forward TD($\lambda$).
This work was partly supported by grants from Alberta Innovates -- Technology Futures and the National Science and Engineering Research Council of Canada.


\newpage
\appendix

\section{Proof TD($\lambda$) is Similar to the Online $\lambda$-Return Algorithm}
\label{sec:proof}

{\bf \noindent Theorem 1}
{\it \, Let $\w_0$ be the initial weight vector, $\w_{t}^{td}$ be the weight vector at time $t$ computed by TD($\lambda$), and $\w_{t}^{\lambda}$ be the weight vector at time $t$ computed by the online $\lambda$-return algorithm.
Furthermore, assume that $\nabla_\w \hat V$ is well-defined and continuous everywhere and that 
$\sum_{i=0}^{t-1} \Delta_i^t \neq {\boldsymbol 0}$.
Then, for all time steps $t$:
$$\frac{|| \w_{t}^{td}  - \w_{t}^{\lambda}  ||}{|| \w_{t}^{td}  - \w_0  ||} \rightarrow 0 \qquad\mbox{as $\,\,\,\alpha \rightarrow 0$}.$$
}

\begin{proof}
We prove the theorem by showing that $|| \w_{t}^{td}  - \w_{t}^{\lambda}  || / || \w_{t}^{td}  - \w_0  ||$ can be approximated by $\mathcal{O}(\alpha) / \big(C +  \mathcal{O}(\alpha)\big)$ as $\alpha \rightarrow 0$,  with $C > 0$. For readability, we will not use the `td' and `$\lambda$' superscripts; instead, we always use weights with double indices for the online $\lambda$-return algorithm and weights with single indices for TD($\lambda$). 

The update equations for TD($\lambda$) are:
\begin{eqnarray*}
\delta_t &=& R_{t+1} + \gamma \hat V(S_{t+1}| \w_t)   - \hat V(S_{t}| \w_t) \\
{\bs e}_t &=& \gamma\lambda {\bs e}_{t-1} +  \nabla_\w  \hat V(S_{t}| \w_t) \\
\w_{t+1} &=&  \w_t + \alpha \delta_t\,{\bs e}_{t}
\end{eqnarray*}
By incremental substitution, we can write $\w_t$ directly in terms of $\w_0$:
\begin{eqnarray*}
{\w}_t &=& \w_0 + \alpha \sum_{j=0}^{t-1} \delta_j {\bs e}_j\\
&=&  \w_0 + \alpha \sum_{j=0}^{t-1} \delta_j \sum_{i=0}^{j} (\gamma\lambda)^{j-i} \,  \nabla_\w  \hat V(S_{i}| \w_i)\\
&=&  \w_0 + \alpha \sum_{j=0}^{t-1} \sum_{i=0}^{j} (\gamma\lambda)^{j-i} \delta_{j}\,  \nabla_\w  \hat V(S_{i}| \w_i)
\end{eqnarray*}
Using the summation rule  $\sum_{j=k}^n \sum_{i=k}^j a_{i,j} = \sum_{i=k}^n \sum_{j=i}^n a_{i,j}$ we can rewrite this as:
\begin{equation}
 {\w}_t = \w_0 + \alpha \sum_{i=0}^{t-1} \sum_{j=i}^{t-1} (\gamma\lambda)^{j-i} \delta_{j}  \nabla_\w \hat V(S_i | {\w}_i) 
 \label{eq:eq1}
 \end{equation}
In Appendix B, the following relation is proven (see Equation \ref{eq:4546}):
$$G^{\lambda | h+1}_i  = G^{\lambda | h}_i   + (\gamma\lambda)^{h-i} \delta'_{h}\qquad\mbox{for } h \geq i+1$$
with
$$\delta'_h := R_{h+1} + \gamma  \hat V(S_{h+1} | \w_{h})  - \hat V(S_{h} | \w_{h-1})  \,.$$
By applying this sequentially for $i+1 \leq h < t$, we can derive:
\begin{equation}
G^{\lambda | t}_i  =   G_i^{\lambda | i+1} + \sum_{j=i+1}^{t-1}   (\gamma\lambda)^{j-i} \delta'_{j}
\label{eq:eq33}
\end{equation}
Furthermore, the following holds:
\begin{eqnarray*}
G_i^{\lambda|i+1} &=& R_{i+1} + \gamma \hat V(S_{i+1}| \w_i) \\
&=& R_{i+1} + \gamma \hat V(S_{i+1}| \w_i) - \hat V(S_{i}| \w_{i-1}) + \hat V(S_{i}| \w_{i-1}) \\
&=& \delta'_i + \hat V(S_{i}| \w_{i-1})  
\end{eqnarray*}
Substituting this in (\ref{eq:eq33}) yields:
$$G^{\lambda | t}_i  =   \hat V(S_{i}| \w_{i-1}) + \sum_{j=i}^{t-1}   (\gamma\lambda)^{j-i} \delta'_{j}\,.$$
Using that $\delta_j' = \delta_j  +  \hat V(S_{j}| \w_{j})  -  \hat V(S_{j}| \w_{j-1})$, it follows that
\begin{equation}
\sum_{j=i}^{t-1}   (\gamma\lambda)^{j-i} \delta_{j} = G^{\lambda | t}_i  -  \hat V(S_{i}| \w_{i-1}) -  \sum_{j=i}^{t-1}   (\gamma\lambda)^{j-i}  \big(\hat V(S_{j}| \w_{j})  -  \hat V(S_{j}| \w_{j-1})\big)\,.
\label{eq:eq345}
\end{equation}
From the update equations of TD($\lambda$) it follows that $|| \w_j - \w_{j-1} || \rightarrow 0 $ as $\alpha \rightarrow 0$. Furthermore, because $\nabla_\w \hat V$ is well-defined everywhere, $\hat V$ is a continuous function, and therefore if $|| \w_j - \w_{j-1} || \rightarrow 0 $ then $||\hat V(S_{j}|\w_{j-1}) - \hat V(S_{j}|\w_{j})|| \rightarrow 0$. Hence, as $\alpha \rightarrow 0$, we can approximate (\ref{eq:eq345}) as:
\begin{eqnarray*}
\sum_{j=i}^{t-1}   (\gamma\lambda)^{j-i} \delta_{j} &=& G^{\lambda | t}_i  -  \hat V(S_{i}| \w_{i-1})+ \mathcal{O}(\alpha)\\
&=& \bar G^{\lambda | t}_i  -  \hat V(S_{i}| \w_{0})+ \mathcal{O}(\alpha)
\end{eqnarray*}
with $\bar G^{\lambda | t}_i$ the interim $\lambda$-return that uses $\w_0$ for all value evaluations. Substituting this in (\ref{eq:eq1}) yields:
$${\w}_t = \w_0 + \alpha \sum_{i=0}^{t-1} \Big( \bar G^{\lambda | t}_i  - \hat V(S_i | {\w}_0) + \mathcal{O}(\alpha)  \Big) \nabla_\w \hat V(S_i | {\w}_i)$$
Because $\nabla_\w \hat V$ is a continuous function, if $|| \w_i - \w_{0} || \rightarrow 0 $ then $||\nabla_\w \hat V(S_{i}|\w_{i}) - \nabla_\w \hat V(S_{i}|\w_{0})|| \rightarrow 0$. Using this, we can approximate the above equation further  as:
\begin{eqnarray}
 {\w}_t &=& \w_0 + \alpha \sum_{i=0}^{t-1} \Big( \bar G^{\lambda | t}_i  - \hat V(S_i | {\w}_0) + \mathcal{O}(\alpha)  \Big) \Big( \nabla_\w \hat V(S_i | {\w}_0) + \mathcal{O}(\alpha) \cdot {\bs 1}\Big) \nonumber \\
 &=& \w_0 + \alpha \sum_{i=0}^{t-1} \Big( \bar G^{\lambda | t}_i  - \hat V(S_i | {\w}_0) \Big) \nabla_\w \hat V(S_i | {\w}_0) + \mathcal{O}(\alpha^2) \cdot {\bs 1}\,,
 \label{eq:9473}
\end{eqnarray}
with ${\bs 1}$ a vector consisting only of 1's.

For the online $\lambda$-return algorithm, we can derive the following by sequential substitution:
$$\w_t^t = \w_0 + \alpha \sum_{i=0}^{t-1} \Big( G_i^{\lambda|t} - \hat V(S_i | \w_i^t) \Big) \nabla_\w \hat V(S_i | \w_i^t)$$
As $\alpha \rightarrow 0$, we can approximate this as:
\begin{equation}
\w_t^t = \w_0 + \alpha \sum_{i=0}^{t-1} \Big( \bar G^{\lambda | t}_i  - \hat V(S_i | {\w}_0) \Big) \nabla_\w \hat V(S_i | {\w}_0) + \mathcal{O}(\alpha^2) \cdot {\bs 1}\,.
\label{eq:eq3423}
\end{equation}

Combining (\ref{eq:9473}) and (\ref{eq:eq3423}), it follows that as $\alpha \rightarrow 0$:
$$\frac{|| \w_{t}  - \w_{t}^{t}  ||}{|| \w_{t}  - \w_0  ||} = \frac{|| (\w_{t}  - \w_{t}^{t})/\alpha  ||}{|| (\w_{t}  - \w_0)/\alpha  || }  = \frac{\mathcal{O}(\alpha)}{C +  \mathcal{O}(\alpha)}\,,$$
with 
$$ C = \left|\left|  \sum_{i=0}^{t-1} \Big( \bar G^{\lambda | t}_i  - \hat V(S_i | {\w}_0) \Big) \nabla_\w \hat V(S_i | {\w}_0)  \right|\right|\ = \left|\left| \sum_{i=0}^{t-1} \Delta_i^t \right|\right|\,.$$
From the condition $\sum_{i=0}^{t-1} \Delta_i^t \neq {\boldsymbol 0}$ it follows that $C > 0$.
\end{proof}

\section{Efficiently Computing the $K$-bounded $\lambda$-Return}
\label{sec:bounded lambda-return}

Here, we derive the two update equations that underly forward TD($\lambda$). First, we derive the equation to compute $G^{\lambda | h+1}_t $ from $G^{\lambda | h}_t$. We use $\hat V_t$ as a shorthand for $\hat V(S_t | \w_{t-1}$).
The value $G^{\lambda | h+1}_t$ can be written in terms of $G^{\lambda | h}_t$ as follows:
\begin{eqnarray}
G^{\lambda | h+1}_t \!\!&:=&  (1-\lambda) \sum_{n=1}^{h-t}  \lambda^{n-1} G_t^{(n)} + \lambda^{h-t} G_t^{(h+1-t)} \nonumber \\
&=& (1-\lambda) \sum_{n=1}^{h-t-1}  \lambda^{n-1} G_t^{(n)} + \lambda^{h-t} G_t^{(h+1-t)}  + \,(1-\lambda) \lambda^{h-t-1} G_t^{(h-t)}  \nonumber \\
&=& (1-\lambda) \sum_{n=1}^{h-t-1}  \lambda^{n-1} G_t^{(n)} +  \lambda^{h-t-1} G_t^{(h-t)} + \lambda^{h-t} \big(G_t^{(h+1-t)}  - G_t^{(h-t)}\big)  \nonumber \\
&=&  G^{\lambda | h}_t   + \lambda^{h-t} \big(G_t^{(h+1-t)}  - G_t^{(h-t)}\big)
\label{eq:part1}
\end{eqnarray}
Furthermore, we can rewrite the difference $G_t^{(h+1-t)}  - G_t^{(h-t)}$ as follows:
\begin{eqnarray*}
G_t^{(h+1-t)}  - G_t^{(h-t)} \!\!\!&=&\!\!\!  \sum_{k=1}^{h+1-t} \gamma^{k-1} R_{t+k} + \gamma^{h+1-t}\, \hat V_{h+1} -   \sum_{k=1}^{h-t} \gamma^{k-1} R_{t+k} - \gamma^{h-t}\, \hat V_h \nonumber \\
&=& \gamma^{h-t} \Big(R_{h+1} + \gamma \hat V_{h+1}  -  \hat V_h \big)\nonumber \\
\end{eqnarray*}
By combining this expression with (\ref{eq:part1}), we get:
\begin{equation}
G^{\lambda | h+1}_t  = G^{\lambda | h}_t   + (\gamma\lambda)^{h-t} \delta'_{h}\,,
\label{eq:4546}
\end{equation}
with
$$\delta'_h := R_{h+1} + \gamma  \hat V(S_{h+1} | \w_{h})  - \hat V(S_{h} | \w_{h-1})  \,.$$

Next, the derive the equation to compute $G_{t+1}^h$ from $G_{t}^h$. The first step in the derivation makes use of the fact that the weights of the $n$-step returns in the $K$-bounded $\lambda$-return always sum to 1. That is, for $0 \leq \lambda \leq 1$ and $n \in \mathbb{N}^+$, the following holds (this can be proven using the geometric series rule):
\begin{equation}
(1-\lambda) \sum_{i=1}^{n-1} \lambda^{i-1}  +  \lambda^{n-1} = 1
\end{equation}
In addition, the derivation makes use of the following relation, for \mbox{$n \geq 2$}:
\begin{eqnarray*}
G^{(n)}_t &=&  \sum_{i=1}^n \gamma^{i-1} R_{t+i} + \gamma^n\, \hat V_{t+n}\\
&=& R_{t+1} + \sum_{i=2}^n \gamma^{i-1} R_{t+i} + \gamma^n\, \hat V_{t+n}\\
&=& R_{t+1} + \sum_{j=1}^{n-1} \gamma^{j} R_{t+1+j} + \gamma^n\, \hat V_{t+n}\\
&=& R_{t+1} + \gamma\Big[\sum_{j=1}^{n-1} \gamma^{j-1} R_{t+1+j} + \gamma^{n-1}\, \hat V_{t+n} \Big]\\
&=& R_{t+1} + \gamma G^{(n-1)}_{t+1} 
\end{eqnarray*}
The full derivation is as follows ($h \geq t+2$):
\begin{eqnarray*}
G^{\lambda | h}_t  \!\!\!\!\!&:=&  (1-\lambda) \sum_{i=1}^{h-t-1} \lambda^{i-1} G^{(i)}_t + \lambda^{h-t-1} G^{(h-t)}_t \\
&=&  (1-\lambda) \sum_{i=1}^{h-t-1} \lambda^{i-1} G^{(i)}_t + \lambda^{h-t-1} G^{(h-t)}_t  + R_{t+1} -  \big[ (1-\lambda) \sum_{i=1}^{h-t-1} \lambda^{i-1}  + \lambda^{h-t-1} \big] R_{t+1}\\
&=&  (1-\lambda) \sum_{i=1}^{h-t-1} \lambda^{i-1} \big[G^{(i)}_t  - R_{t+1} \big]  + \lambda^{h-t-1} \big[ G^{(h-t)}_t - R_{t+1} \big]  + R_{t+1}\\
&=&  (1-\lambda) \big[G^{(1)}_t  - R_{t+1} \big] + \, (1-\lambda) \sum_{i=2}^{h-t-1} \lambda^{i-1} \big[G^{(i)}_t  - R_{t+1} \big]\\
&& +\, \lambda^{h-t-1} \big[ G^{(h-t)}_t - R_{t+1} \big]  + R_{t+1}\\
&=&  (1-\lambda) \big[R_{t+1} + \gamma \hat V_{t+1}  - R_{t+1} \big]  + \,(1-\lambda) \sum_{i=2}^{h-t-1} \lambda^{i-1} \gamma G^{(i-1)}_{t+1}\\
&& + \,\lambda^{h-t-1} \gamma G^{(h-t-1)}_{t+1}  + R_{t+1}\\
&=&  \gamma(1-\lambda)  \hat V_{t+1} +\, (1-\lambda) \sum_{j=1}^{h-t-2} \lambda^{j} \gamma G^{(j)}_{t+1} +\, \lambda^{h-t-1} \gamma G^{(h-t-1)}_{t+1}  + R_{t+1}\\
&=&  \gamma(1-\lambda)  \hat V_{t+1} + R_{t+1}  + \,\gamma\lambda \Big[ (1-\lambda) \sum_{j=1}^{h-t-2} \lambda^{j-1} G^{(j)}_{t+1} +  \lambda^{h-t-2} G^{(h-t-1)}_{t+1} \Big]\\
&=&  \gamma(1-\lambda)  \hat V_{t+1} + R_{t+1} + \gamma\lambda G_{t+1}^{\lambda | h} \\
\end{eqnarray*}
The above derivation expresses $G_t^{\lambda|h}$ in terms of  $G_{t+1}^{\lambda | h}$. $G_{t+1}^{\lambda | h}$ expressed in terms of $G_t^{\lambda|h} $ yields:
$$G^{\lambda | h}_{t+1} = (G^{\lambda | h}_t - \rho_t )/\gamma\lambda\,,  \qquad\mbox{ for } h \geq t+2$$
with
$$ \rho_t = R_{t+1} + \gamma(1-\lambda)\, \hat V(S_{t+1}| \w_t)\,.$$

\end{document}